\documentclass[11pt,a4paper]{article}
\usepackage[hyperref]{emnlp2018}
\usepackage{times}
\usepackage{latexsym}
\usepackage{url}

\usepackage[]{graphicx}
\usepackage[tbtags]{amsmath}
\usepackage{multirow}

\aclfinalcopy 

\title{Learning End-to-End Goal-Oriented Dialog with Multiple Answers}

\author{
  Janarthanan Rajendran\thanks{\hspace{0.2cm}Equal Contribution}\\
  University of Michigan \\
  {\tt rjana@umich.edu}
  \\\And
  Jatin Ganhotra\footnotemark[1] \\
  IBM Research \\
  {\tt jatinganhotra@us.ibm.com}
  \\\AND
  Satinder Singh \\
  University of Michigan \\
  {\tt baveja@umich.edu}
  \\\And
  Lazaros Polymenakos \\
  IBM Research \\
  {\tt lcpolyme@us.ibm.com}
  }

\date{}

\begin{document}
\maketitle
\begin{abstract}
In a dialog, there can be multiple valid next utterances at any point. The present end-to-end neural methods for dialog
do not take this into account. They learn with the assumption that at any time there is only one correct next utterance.
In this work, we focus on this problem in the goal-oriented dialog setting where there are different paths to reach a goal.
We propose a new method, that uses a combination of supervised learning and reinforcement learning approaches to address this issue. We also propose a new and more effective testbed, permuted-bAbI dialog tasks
\footnote{permuted-bAbI-dialog-tasks - \url{https://github.com/IBM/permuted-bAbI-dialog-tasks}}
by introducing multiple valid next utterances to the original-bAbI dialog tasks, which allows evaluation of goal-oriented dialog systems in a more realistic setting. We show that there is a significant drop in performance of existing end-to-end neural methods from 81.5\% per-dialog accuracy on original-bAbI dialog tasks to 30.3\% on permuted-bAbI dialog tasks. We also show that our proposed method improves the performance and achieves 47.3\% per-dialog accuracy on permuted-bAbI dialog tasks.

\end{abstract}

\section{Introduction}
\label{introduction}

End-to-end, neural conversation models that learn from chatlogs of human-to-human interaction hold the promise of quickly bootstrapping dialog systems and keep evolving them based on new data. Recent work (\cite{vinyals2015neural, bordes2016learning, serban2016building}) has shown that dialog models can be trained in an end-to-end manner with satisfactory results. 

However, human dialog has some unique properties that many other learning tasks do not. For any given dialog state or input,  multiple correct next utterances or answers may be possible; i.e given the dialog so far, there are several different utterances which one can say next that would be valid. However, this property of dialog is not taken into account in the present way of training end-to-end neural dialog systems. 

There are two broad ways in which present day neural dialog systems can be trained: Supervised Learning (SL) and Reinforcement Learning (RL). In the RL setting, the dialog system learns through trial and error with reinforcement (rewards at the end or at key dialog points) from a human or a simulator. RL training for dialog is a hard problem to solve.
It is difficult to define and award appropriate rewards, and to learn language from scratch through these rewards.
RL training also demands a large amount of training interaction. In order to handle these challenges, RL methods are almost always complimented with a SL phase.

In SL setting, a fixed set of dialog data is collected from humans and the dialog system is trained to imitate that data. When a new dialog dataset is curated, the data is extracted from real-world chat logs from human-human conversation, where one human acts as the agent. It is not possible to know all of the valid next utterances for a given dialog state at any single time. A particular dialog in the dataset has access to only one of the valid next utterances given the dialog history and the current utterance. Another valid next utterance could be present in some other dialog in the dataset.

Since for a given dialog only one correct answer is available at any single time, the gradients are calculated based on the assumption that there is only one correct next utterance for the given dialog state. This results in reducing the probability of other valid next utterances for that dialog. While all this is true for dialog in general, in this work, we focus on the goal-oriented dialog setting.

We propose a novel method which handles the issue of learning multiple possibilities for completing a goal-oriented dialog task. Our presentation is organized as follows: In Sections 2 and 3, we define the multiple-utterance problem and point out the limitations of current learning methodologies. 

Section 4 describes our proposed method, which combines Supervised Learning and Reinforcement Learning approaches for handling multiple correct next utterances. In Section 5, we introduce permuted-bAbI dialog tasks, which is our proposed testbed for goal-oriented dialog. Section 6 details our experimental results across all datasets and all models.

\section{Multiple-utterance problem in goal-oriented dialog}

Goal-oriented dialog tasks are those in which there is an explicit goal that the system tries to achieve through the dialog. These tasks typically involve getting some information from the user, interacting with an external Knowledge Base (KB) and giving back information to the user. Simple examples (form filling) include restaurant reservation, hotel booking etc., whereas complex tasks could involve a combination of informative and form filling tasks (e.g. IT support, customer care etc.). 

There could be multiple ways/strategies to achieve a given task. When a dataset is collected from different people performing the task, these different ways of solving the task get reflected in it. These variations could be as simple as difference in the order in which the system asks the information from the user, or as complex as following a completely different line of questions/answers to achieve the task. For example, in an IT support scenario, one may ask a sequence of standard questions or start from common problems and once eliminated, follow the standard set of questions. In the dataset, they turn into multiple valid next utterances for a given dialog so far.

Our objective in this paper is not to mimic all humans from whom the data was collected from or all the possible strategies, but rather to use that knowledge and learn to perform the task better and faster.  

\section{Issues with the present methods}
Consider a goal-oriented dialog dataset for restaurant reservation where the dialog system has to acquire cuisine, location, number of people and price range information from the user before retrieving restaurant options. Consider two dialogs (A and B) in the dataset which have the same first system utterance (\texttt{S1a} is same as \texttt{S1b}). Let their dialog state vector after encoding the dialog until \texttt{S1} be $s$. This state vector $s$ is what will be used for next utterance generation or retrieval. Their next utterance is different because of the variation in the order in which the information is asked from the user as shown below. i.e \texttt{S2a} is different from \texttt{S2b}.\\

{\small
\noindent \textbf{Dialog A}\\
U: Hello !\\
\texttt{S1a: Hello, what can I help you with today?}\\
U: Can you book a table for six people with French food?\\
\texttt{S2a: Where should it be?}\\
U: In Bombay\\
\texttt{S3a: Which price range are you looking for?}\\
U: I am looking for a cheap restaurant.\\
\texttt{S4a: Ok, let me look into some options for you.}\\
}

{\small
\noindent \textbf{Dialog B}\\
U: Hello !\\
\texttt{S1b: Hello, what can I help you with today?}\\
U: Can you book a table for six people with French food?\\
\texttt{S2b: Which price range are you looking for?}\\
U: I am looking for a cheap restaurant.\\
\texttt{S3b: Where should it be?}\\
U: In Bombay\\
\texttt{S4b: Ok, let me look into some options for you.}\\
}

\begin{figure*}[htbp!]
\centering
\includegraphics[width=1\textwidth]{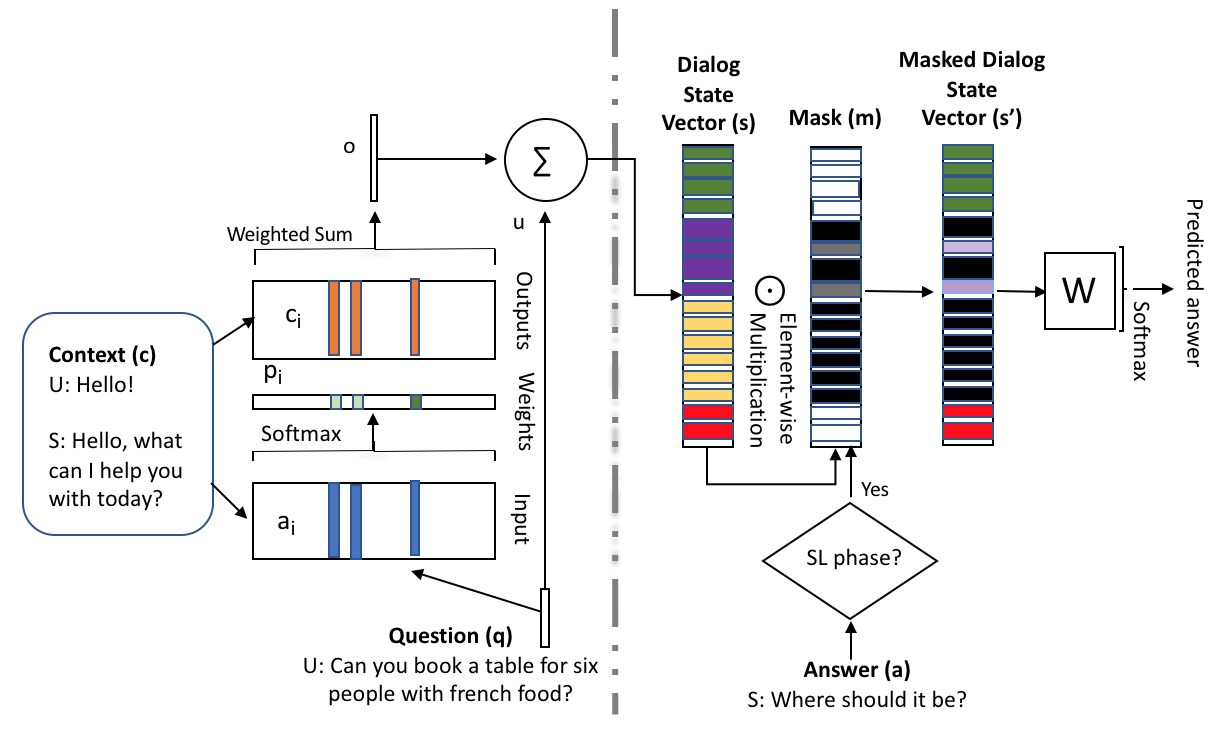}
\caption{\textbf{Mask-memN2N} - \textit{Left}: A single layer version of memN2N. \textit{Right}: Masking} 
\label{fig_idea}
\end{figure*}

These two dialogs might be present in different places in the dataset. When dialog A is part of the batch for which loss is calculated and parameters are updated, the dialog system is asked to produce \texttt{S2a} from $s$. Here, the loss could be negative log-likelihood, squared error or anything that tries to push $s$ towards producing \texttt{S2a}. In this process, the probability of the dialog system producing \texttt{S2b}, an equally valid answer, is reduced. The reverse happens when the dialog system encounters a batch with dialog B. 
This is true whether a softmax or a sigmoid non-linearity for each unit is used in the output layer. In essence, the system is expected to learn a one to many function, but is forced to produce only one of the valid outputs at one time (that is all we have at any one time), while telling that all other outputs are wrong.

Note that this could be a problem even when the dialogs are similar (semantically) in the beginning, but not the same exact dialog. 
For simplicity, we show an example where two dialogs have the same beginning and only 2 valid next utterances occur.

\section{Proposed Method}
\label{proposed_solution}
The proposed method has two phases. In one phase, the dialog system tries to learn how to perform dialog from the dataset by trying to mimic it and in the other it learns through trial and error. The former uses supervised learning and the latter uses reinforcement learning.
Consider a dialog state vector $s$. This has all the information from the dialog so far and is used for next utterance generation or retrieval. 
Any neural method such as memory network \cite{weston2014memory}, HRED \cite{sordoni2015hierarchical} etc. can be used for encoding and producing the dialog state vector $s$. As discussed earlier for the state vector $s$, there could be multiple valid next utterances. 

During the SL phase, at each data point, the dialog system is trained to produce the one next utterance provided in that data point and is penalized even if it produces one of the other valid next utterances. We avoid this by providing the dialog system, the ability to use only parts of the state vector to produce that particular next utterance. 
This allows only parts of the network to be affected that were responsible for the prediction of that particular answer.
The dialog system can retain other parts of the state vector and values in the network that stored information about other valid next utterances.
This is achieved by generating a mask vector $m$ which decides which parts of the state vector $s$ should be used for producing that particular answer. This is achievable, as $m$ is learned as function of $s$ and the actual answer $a$ present in the given dialog data point. 

In the RL phase, however, the dialog system is rewarded if it produces an answer that is among any of the set of valid correct answers. While in the SL phase the dialog system had access to the actual answer $a$ at given time to produce the mask, in the RL phase the dialog system produces the mask by only using the dialog state vector $s$.

\begin{align}
&\textrm{Supervised Learning phase}\\ \nonumber
    &m = \sigma(W_s s + W_a a + b_{sl})\\ \nonumber
    &s'= m * s\\
    &\textrm{Reinforcement Learning phase}\\ \nonumber
    &m = \sigma(W_s s + W_r s + b_{rl})\\ \nonumber
    &s'= m * s
\end{align}

where $W$'s and $b$'s are the parameters learned, $\sigma$ is an element wise sigmoid non-linearity and $s'$ is the masked dialog state vector that is used by the dialog system to perform the downstream task such as next utterance generation or retrieval. The parameters of the network that produce $s$ and that follow $s'$ are shared between the two phases. 
While there are different ways of combining the two phases during training, the RL phase which does not use the answer for its mask is what is used during testing. The masking approach described above is illustrated in Fig \ref{fig_idea}.

In this work, SL phase is performed first, followed by RL phase. In the SL phase, the dialog system learns different dialog responses and behaviours from the dataset. It has the ability to learn multiple possible next utterances without one contradicting/hindering the learning of the other much. In the RL phase, the dialog system might settle on a unique behaviour that it finds best for it to perform the task and uses that during test time as well.

\section{Permuted bAbI dialog tasks}
\label{permuted-babi}

\begin{figure*}[ht]
\centering
\includegraphics[width=1\textwidth]{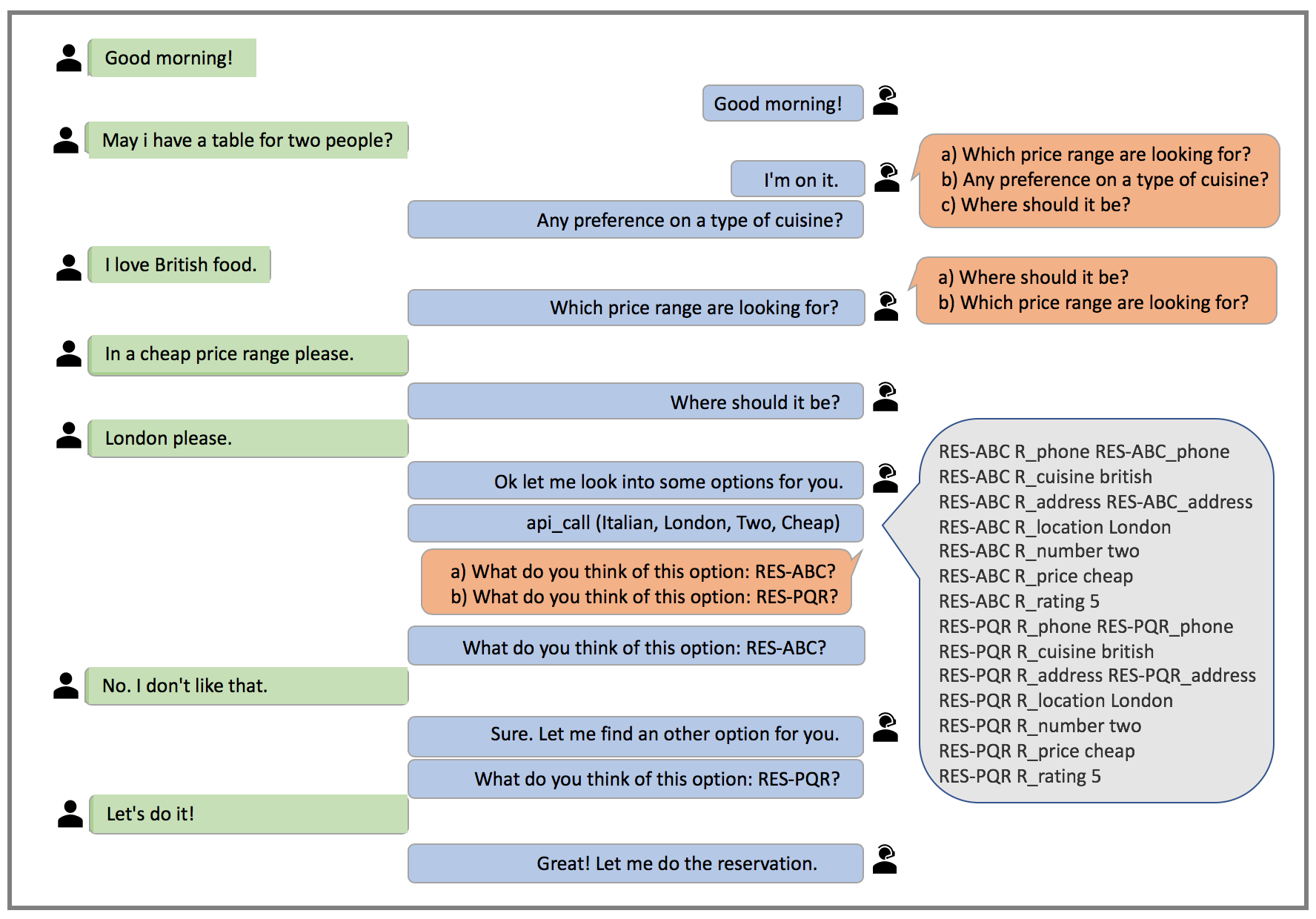}
\caption{ \textbf{Permuted-bAbI dialog tasks.} A user (in green) chats with a dialog system (in blue) to book a table at a restaurant. At a given point in the dialog, the dialog system has multiple correct next utterances (in orange). The dialog system can choose either of the multiple correct utterances as the next utterance. The list of restaurants are returned from the API\_call (in grey) also contain multiple restaurants with the same rating, giving the dialog system more options to propose to the user.} 
\label{fig_dataset}
\end{figure*}

\citet{bordes2016learning} proposed bAbI dialog tasks as a testbed to break down the strengths and shortcomings of end-to-end dialog systems in goal-oriented applications. There are five tasks, generated by a simulation set in the context of restaurant reservation, with the final goal of booking a table. The simulation is based on an underlying KB, whose facts contain restaurants and their properties. 
Tasks 1 (Issuing API calls) and 2 (Updating API calls) test the dialog system to implicitly track dialog state, whereas Task 3 (Displaying options) and 4 (Providing extra information) check if the system can learn to use KB facts in a dialog setting. Task 5 (Conducting full dialogs) combines all tasks.

\cite{bordes2016learning} used natural language patterns to create user and system utterances. There are 43 patterns for the user and 20 for the system, which were combined with the KB entities to form thousands of different utterances. However, on a closer analysis of the testbed, we observe that even though there are thousands of different utterances, these utterances always follow a fixed deterministic order (predefined by the simulation), which makes the tasks easier and unsuitable to mimic conversations in the real-world. For example, for Task 1, the system follows a predefined order to ask for missing fields required to complete the dialog state. In Task 3, all restaurants retrieved have a unique and different rating. While this makes evaluation deterministic and easier, these hidden settings in the simulation create conversations, which are simpler compared to real-world conversations for restaurant reservation.

We propose permuted-bAbI dialog tasks, an extension of original-bAbI dialog tasks, which make our proposed testbed more appropriate for evaluating dialog systems in goal-oriented setting. In \textit{original-bAbI dialog tasks} at a given time in the conversation, there is \textbf{only one} correct system utterance. \textit{Permuted-bAbI dialog tasks} allow \textbf{multiple} correct system utterances at a given point in the conversation. 

We propose the following changes to original-bAbI dialog tasks. In Task 1, a user request defines a query that can contain from 0 to 4 of the required fields to make a reservation. The system asks questions to fill the missing fields and eventually generate the correct corresponding API call. However, the system asks for information in a \textit{deterministic} order -
\\
\textit{Cuisine $\rightarrow$ Location $\rightarrow$ People $\rightarrow$ Price}
\\
to complete the missing fields. In permuted-bAbI dialog tasks, we don't follow a deterministic order and allow 
the system to ask for the missing fields in any order.

In Task 3, for the API call matching the user request, the facts are retrieved from the KB and provided as part of dialog history. The system must propose options to users by listing the restaurant names sorted by their corresponding rating (from higher to lower) until users accept. However, each restaurant has a different rating. In permuted-bAbI dialog tasks, multiple restaurants can have the same rating but the system must still propose the restaurant names following the decreasing order of rating, which allows multiple valid next utterances.

In Task 5, Tasks 1-4 are combined to generate full dialogs.
In permuted-bAbI dialog tasks, we incorporate the changes for both Task 1 and Task 3 mentioned above to the final Task 5 (Conducting full dialogs).

Fig \ref{fig_dataset} shows a dialog sample from permuted-bAbI dialog tasks. Our proposed testbed (We release and show experiments on permuted version of Task 5, i.e. Conducting full dialogs set, as tasks 1-4 are subsets of a full conversation and don't represent a complete meaningful conversation standalone) is more closer to a real-world restaurant reservation conversation, in comparison to the original-bAbI dialog tasks. We release two versions of permuted-bAbI dialog tasks - permuted-bAbI dialog task\text{*} (the full dataset with all permutations. There are around 11,000 dialogs in each set and the exact number varies for train, val, test and test-OOV sets), which contains all permutations (the full dataset) and permuted-bAbI dialog task, which contains 1000 dialogs randomly sampled (we used random seed = 599 for sampling) from permuted-bAbI dialog task\text{*}. We choose a random 1000 subset from each of train, val, test and test-OOV sets to match the number of dialogs in original-bAbI dialog task. Another key point to choose a small subset and to not include all permutations in the training set is that it allows to mimic real-world data collection. For a real-world use-case, as the number of required fields and user options increase, the cost for gathering data covering all permutations will increase exponentially, and one can't guarantee that enough training examples for all permutations will be present in the collected dataset. Note that, since there are multiple correct next utterances, we also modify the evaluation criteria so that the system is rewarded if it predicts any of the multiple correct next utterances.


\section{Experiments and Results}
\label{experiments_and_results}

\begin{table*}[htbp!]
\begin{center}
\begin{tabular}{lcc|cc}
    \hline
    \multirow{2}{*}{\textbf{Dataset}} & \multicolumn{2}{c}{\textbf{no match-type}} & \multicolumn{2}{c}{\textbf{+ match-type}}\\
     & \textbf{Per-turn} & \textbf{Per-dialog} & \textbf{Per-turn} & \textbf{Per-dialog} \\ \hline
    Original-bAbI dialog task & 98.5 & 77.1 & 98.8 & 81.5 \\ \hline
    Permuted-bAbI dialog task\text{*} & 96.4 & 58.2 & 96.9 & 63.9\\ \hline
    Permuted-bAbI dialog task & 91.8 & 22 & 93.3 & 30.3\\ \hline
    \hline
    OOV: Original-bAbI dialog task & 65.6 & 0  & 78.3 & 0 \\ \hline
    OOV: Permuted-bAbI dialog task\text{*} & 63.6 & 0 & 78.4 & 0  \\ \hline
    OOV: Permuted-bAbI dialog task & 63.4 & 0.5 & 78.1 & 0.6 \\ \hline
\end{tabular}
\end{center}
\caption{\textbf{Test results for our baseline end-to-end memory network model across the three datasets.} Results (accuracy \%) are given in the standard setup and out-of-vocabulary (OOV) setup. Results are given for both with and without match-type features.}
\label{tab:baseline-results}
\end{table*}

End-to-end memory networks \cite{sukhbaatar2015end} are an extension of Memory Networks proposed by \cite{weston2014memory} which have been successful on various natural language processing tasks. End-to-end memory networks are trained end-to-end and use a memory component to store dialog history and short-term context to predict the required response. They perform well on original-bAbI dialog tasks and have been shown to outperform some other end-to-end architectures based on Recurrent Neural Networks. Hence, we chose them as end-to-end model baseline. We perform experiments on the three datasets mentioned above across all our models. We also perform experiments with match-type features proposed by \cite{bordes2016learning-memN2N}, which allow the model to use type-information for entities like location, cuisine, phone number etc. The results for our baseline model, our proposed model and results on our ablation study are described below. The test results reported are calculated by choosing the model with highest validation per-turn accuracy across multiple runs.

\begin{table*}[htbp!]
\begin{center}
\begin{tabular}{lcc|cc}
    \hline
    \multirow{2}{*}{\textbf{Model}} & \multicolumn{2}{c}{\textbf{no match-type}} & \multicolumn{2}{c}{\textbf{+ match-type}}\\
     & \textbf{Per-turn} & \textbf{Per-dialog} & \textbf{Per-turn} & 
     \textbf{Per-dialog} \\ \hline
    memN2N &  91.8 & 22 & 93.3 & 30.3\\ \hline
    memN2N + all-answers$\dagger$ & 88.5 & 14.9 & 92.5 & 26.4\\ \hline
    Mask-memN2N &  93.4 & \textbf{32} & 95.2 & \textbf{47.3}\\\hline
    \hline
    OOV: memN2N & 63.4 & 0.5 & 78.1 & 0.6  \\ \hline
    OOV: memN2N + all-answers$\dagger$  & 60.8 & 0.5 & 74.9 & 0.6 \\ \hline
    OOV: Mask-memN2N & 63.0  & 0.5 & 80.1 & 1 \\\hline
\end{tabular}
\end{center}
\caption{\textbf{Test results for various models on permuted-bAbI dialog task.} Results (accuracy \%) are given in the standard setup and OOV setup; and both with and without match-type features.}
\label{tab:our-model-results}
\end{table*}

\subsection{Baseline model: memN2N}
\label{baseline-model-results}
A single layer version of the memN2N model is shown in Fig.\ref{fig_idea}. A given sentence $(i)$ from the context (dialog history) is stored in the memory by it's input representation $(a_{i})$. Each sentence $(i)$ also has a corresponding output representation $(c_{i})$. To identify the relevance of a memory for the next-utterance prediction, attention of query over memory is computed via dot product, where $(p_{i})$ represents the probability for each memory in equation \ref{eq3}. An output vector $(o)$ is computed by a weighted sum of the memory embeddings $(c_{i})$ with their corresponding probabilities in equation \ref{eq4}. The output vector $(o)$ represents the overall embedding for the context. The output vector $(o)$ and query $(u)$ added together represent the dialog state vector $(s)$ in equation \ref{eq5}.
\begin{align}
\label{eq3}
p_{i} &= \textrm{Softmax}(u^{T} (a_{i})) \\
\label{eq4}
o &= \sum_{i}p_{i}c_{i} \\
\label{eq5}
s &= (o + u)
\end{align}

Our results for our baseline model across the three datasets are given in Table \ref{tab:baseline-results}.
The hyperparameters used for training the baseline models are provided in Appendix \ref{appendix_a1}.

The first 3 rows show the results for the three datasets in the standard setup, and rows 4-6 show results in the Out-Of-Vocabulary (OOV) setting. Per-response accuracy counts the percentage of responses that are correct (i.e., the correct candidate is chosen out of all possible candidates). Note that, as mentioned above in Section \ref{permuted-babi}, since there are multiple correct next utterances, a response is considered correct if it predicts any of the multiple correct next utterances. Per-dialog accuracy counts the percentage of dialogs where every response is correct. Therefore, even if only one response is incorrect, this might result in a failed dialog, i.e. failure to achieve the goal of restaurant reservation. We report results both with and without match type features, shown in the last two columns.

From Table \ref{tab:baseline-results}, we observe that the baseline model performs poorly on permuted-bAbI dialog tasks (both full dataset and 1000 random dialogs). For permuted-bAbI dialog task\text{*}, the baseline model achieves 58.2\% on per-dialog accuracy, but the number decreases to only 22\% for permuted-bAbI dialog task (1000 random dialogs). This implies that only 1 out of every 4 dialogs might be successful in completing the goal. The results improve slightly by using match-type features, but 30\% per-dialog accuracy is still very low for real-world use. These results clearly demonstrate that the end-to-end memory network model does not perform well on our proposed testbed, which is more realistic and mimics real-world conversations more closely.

\subsection{Mask End-to-End Memory Network: Mask-memN2N}
Our model, Mask-memN2N, shown in Fig \ref{fig_idea}, is built on the baseline memN2N model described above, except for an additional masking performed to the dialog state vector. The SL phase is performed for the first 150 epochs. The best performing model chosen based on validation accuracy is used as a starting point for the RL phase. All parameters except for the network that produces the masks are shared between the two phases. During the SL phase, the mask parameters of the RL phase are pre-trained to match the mask produced in SL phase using an L2 loss. Through this approach, when the model transitions in the RL phase, it does not need to explore the valid masks and hence, the answers from scratch. Instead, its exploration will now be more biased towards relevant answers. For the RL phase, we use REINFORCE \cite{williams1992simple} for training the system. An additional loss term is added to increase entropy. The hyperparameters used, including the exact reward function are provided in Appendix \ref{appendix_a2}.

\begin{table*}[htbp!]
    \centering
    \begin{tabular}{l c c}
    \hline
        \textbf{Model} & \textbf{Per-turn} & \textbf{Per-dialog}\\ \hline
        Mask-memN2N &  93.4 & 32\\\hline
        Mask-memN2N (w/o entropy) & 92.1 & 24.6\\\hline
        Mask-memN2N (w/o L2 mask pre-training) & 85.8 & 2.2 \\\hline
        Mask-memN2N (Reinforcement learning phase only) & 16.0 & 0 \\ \hline
    \end{tabular}
    \caption{\textbf{Ablation study of our proposed model on permuted-bAbI dialog task.} Results (accuracy \%) are given in the standard setup, without match-type features.}
    \label{tab:ablation-study}
\end{table*}

\subsection{Model comparison}
\label{proposed-model-results}
Our results for our proposed model and comparison with other models for permuted-bAbI dialog task are given in Table \ref{tab:our-model-results}. Table \ref{tab:our-model-results} follows the same format as Table \ref{tab:baseline-results}, except we show results for different models on permuted-bAbI dialog task. We show results for three models - memN2N, memN2N + all-answers and our proposed model, Mask-memN2N.

In the memN2N + all-answers model, we extend the baseline memN2N model and though not realistic, we provide information on all correct next utterances during training, instead of providing only one correct next utterance. 
The memN2N + all-answers model has an element-wise sigmoid at the output layer instead of a softmax, allowing it to predict multiple correct answers.
This model serves as an important additional baseline, and clearly demonstrates the benefit of our proposed approach.

From Table \ref{tab:our-model-results}, we observe that the memN2N + all-answers model performs poorly, in comparison to the memN2N baseline model both in standard setup and OOV setting, as well as with and without match-type features. This shows that the existing methods do not improve the accuracy of a dialog system even if all correct next utterances are known and used during training the model. 
Our proposed model performs better than both the baseline models. In the standard setup, the per-dialog accuracy increases from 22\% to 32\%. Using match-type features, the per-dialog accuracy increases considerably from 30.3\% to 47.3\%. In the OOV setting, all models perform poorly and achieve per-dialog accuracy of 0-1\% both with and without match-type features. These results are similar to results for original-bAbI dialog Task 5 from \citet{bordes2016learning-memN2N} and our results with the baseline model.

Overall, Mask-memN2N is able to handle multiple correct next utterances present in permuted-bAbI dialog task better than the baseline models. This indicates that permuted-bAbI dialog task is a better and effective evaluation proxy compared to original-bAbI dialog task for real-world data. This also shows that we need better neural approaches, similar to our proposed model, Mask-memN2N, for goal-oriented dialog in addition to better testbeds for benchmarking goal-oriented dialogs systems.

\subsection{Ablation study}
\label{ablation-study}

Here, we study the different parts of our model for better understanding of how the different parts influence the overall model performance. 
Our results for ablation study are given in Table \ref{tab:ablation-study}.
We show results for Mask-memN2N in various settings - a) without entropy, b) without pre-training mask c) reinforcement learning phase only. 

Adding entropy for the RL phase seems to have improved performance a bit by assisting better exploration in the RL phase. When we remove L2 mask pre-training, there is a huge drop in performance. In the RL phase, the action space is large. In the bAbI dialog task, which is a retrieval task, it is all the candidate answers that can be retrieved which forms the action set. L2 mask pre-training would help the RL phase to try more relevant actions from the very start.

From Table \ref{tab:ablation-study} it is clear that the RL phase individually does not perform well; it is the combination of both the phases that gives the best performance. 
When we do only the RL phase, it might be very tough for the system to learn everything by trial and error, especially because the action space is so large. Preceding it with the SL phase and L2 mask pre-training would have put the system and its parameters at a good spot from which the RL phase can improve performance. 
Note that it would not be valid to check performance of the SL phase in the test set as the SL phase requires the actual answers for it to create the mask.

\section{Related Work}
\label{related_works}
End-to-end dialog systems have been trained to show satisfactory performance in goal-oriented setting, as shown by \cite{bordes2016learning} and \cite{seo2016query}).
The idea of allowing the system to learn to attend to different parts of the state vector at different times depending on the input that the proposed model uses has been used in different settings before. To name a few, \citet{bahdanau+al-2014-nmt} use it for Machine Translation (MT), where the MT system can attend to different words in the input language sentence while producing different words in the output language sentence. \citet{pmlr-v37-xuc15} use it for image caption generation where the system attends to different parts of the image while generating different words in the caption. \citet{DBLP:journals/corr/MadottoA17} use it for Question Answering (QA), where the QA system attends and updates different parts of the Recurrent Neural Network story state vector based on the sentence the system is reading in the input story.

In the past, goal-oriented dialog systems would model the conversation as partially observable Markov decision processes (POMDP) (\citet{young2013pomdp}). However, such systems require many hand-crafted features for the state and action space representations, which limited their use only to narrow domains. 
In recent years, several corpora have been made available for building data-driven dialog systems \cite{serban2015survey}. However, there are no good resources to train and test end-to-end models in goal-oriented scenarios. 

Some datasets are proprietary (e.g., \citet{chen2016end}) or require participation to a specific challenge and signing a license agreement (e.g., DSTC4 \cite{kim2017fourth}). Several datasets have been designed to train or test dialog state tracker components (\citet{henderson2014word}, \citet{asri2017frames}, which are unsuitable for training end-to-end dialog systems, either due to limited number of conversations or due to noise.
Recently, some datasets have been designed using crowdsourcing (\citet{hixon2015learning}, \citet{wen2015semantically}, \citet{su2015learning}) e.g., Amazon Mechanical Turk, CrowdFlower etc., but dialog systems built for them are harder to test automatically and involve another set of crowdsource workers for comparing them.

\cite{bordes2016learning} proposed goal-oriented bAbI dialog tasks, a testbed created from a simulation of restaurant reservation setting, which allows easy evaluation and interpretation of end-to-end dialog systems. However, the testbed doesn't mimic real world conversations and conversations are generated around a deterministic set of patterns. Recently, datasets designed using Wizard-of-Oz seting (\citet{eric2017key}, \citet{wen2016network}) show promise, but they are limited in scale, and evaluation metrics are either based around BLEU score and entity F1 scores or require crowdsource workers for evaluation.

\section{Conclusion}
\label{conclusion}

We propose a method that uses masking to handle the issue of making wrong updates at different times because of the presence of multiple valid next utterances in a dataset, but having access to only one of them at any time. The method has a SL phase where the mask uses the answer as well, and an RL phase, where the system learns to generate the mask solely from its dialog state vector. 

We modify the synthetic original-bAbI dialog task to create a more realistic and effective testbed, permuted-bAbI dialog task (which we have made publicly available), that has the issue of multiple next utterances as would be the case with any dataset created from human-human dialogs. 
Our experiments show that there is a significant drop in performance of the present neural methods in the permuted-bAbI dialog task compared to the original-bAbI dialog task. The experiments also confirm that the proposed method is a step in the right direction for bridging this gap.

\appendix
\section{Appendix: Training Details}
\label{appendix_a}
\subsection{Baseline model: memN2N}
\label{appendix_a1}
\label{sec:baseline-model-hyperparameters}
The hyperparameters used for the baseline model are as follows: hops = 3, embedding\_size = 20, batch\_size = 32. The entire model is trained using stochastic gradient descent (SGD) (learning rate = 0.01) with annealing (anneal\_ratio = 0.5, anneal\_period = 25), minimizing a standard cross-entropy loss between $\hat{a}$ and the true label a. We add temporal features to encode information about the speaker for the given utterance \cite{bordes_e2e_dialog} and use position encoding for encoding the position of words in the sentence \cite{sukhbaatar2015end}. We learn two embedding matrices A and C for encoding story, separate embedding matrix B for encoding query and weight matrices TA and TC for encoding temporal features. The same weight matrices are used for 3 hops\footnote{We used 599 as the random seed for both tf.set\_random\_seed and tf.random\_normal\_initializer for our embedding matrices.}.

\subsection{Mask End-to-End Memory Network: Mask-memN2N}
\label{appendix_a2}

We use the same hyperparameters as the baseline model mentioned above. The additional hyperparameters are as follows: L2 loss coefficient = 0.1 for pre-training the RL phase mask, entropy with linear decay from 0.00001 to 0, positive reward = 5 for every correct action and negative reward = 0.5 for an incorrect action.

\newpage

\bibliography{emnlp2018}
\bibliographystyle{acl_natbib_nourl}

\end{document}